# A Hybrid Both Filter and Wrapper Feature Selection Method for Microarray Classification

Li-Yeh Chuang, Chao-Hsuan Ke, and Cheng-Hong Yang, Member, *IAENG*

*Abstract*—Gene expression data is widely used in disease analysis and cancer diagnosis. However, since gene expression data could contain thousands of genes simultaneously, successful microarray classification is rather difficult. Feature selection is an important pre-treatment for any classification process. Selecting a useful gene subset as a classifier not only decreases the computational time and cost, but also increases classification accuracy. In this study, we applied the information gain method as a filter approach, and an improved binary particle swarm optimization as a wrapper approach to implement feature selection; selected gene subsets were used to evaluate the performance of classification. Experimental results show that by employing the proposed method fewer gene subsets needed to be selected and better classification accuracy could be obtained.

## I. INTRODUCTION

DNA microarray technology allows simultaneous monitoring and measuring of thousands of gene expression activation levels in a single experiment. This technology is currently used in medical diagnosis and gene analysis. Many microarray research projects focus on clustering analysis and classification accuracy. In clustering analysis, the purpose of clustering is to analyze the gene groups that show a correlated pattern of the gene expression data and provide insight into gene interactions and function. Research on classification accuracy is aimed at building an efficient model for predicting the class membership of data, produce a correct label on training data, and predict the label for any unknown data correctly.

Typically, gene expression data possess a high dimension and a small sample size, which makes testing and training of general classification methods difficult. In general, only a relatively small number of gene expression data out of the total number of genes investigated shows a significant correlation with a certain phenotype. In other words, even though thousands of genes are usually investigated, only a very small number of these genes show a correlation with the phenotype in question. Thus, in order to analyze gene expression profiles correctly, feature selection (also called gene selection) is crucial for the classification process.

Methods used for data reduction, or more specifically for feature selection in the context of microarray data analysis, can be classified into two major groups: filter and wrapper model approaches [9].

In the filter model approach a filtering process precedes the actual the classification process. For each feature a weight value is calculated, and features with better weight values are chosen to represent the original data set. However, the filter approach does not account for interactions between features. The wrapper model approach depends on feature addition or deletion to compose subset features, and uses evaluation function with a learning algorithm to estimate the subset features. This kind of approach is similar to an optimal algorithm that searches for optimal results in a dimension space. The wrapper approach usually conducts a subset search with the optimal algorithm, and then a classification algorithm is used to evaluate the subset.

Particle swarm optimization (PSO) is a population-based stochastic optimization technique, which was developed by Kennedy and Eberhart in 1995 [3]. PSO simulates the social behavior of organisms, such as birds in a flock or fish in a school, and can be described as an automatically evolving system. In PSO, each single candidate solution can be considered "an individual bird in the flock", that is, a particle in the search space. Each particle makes use of its own memory and knowledge gained by the swarm as a whole to find the best (optimal) solution. In 1997, Kennedy and Eberhart introduced a binary version of PSO (BPSO) [4] to solve discrete problems. In BPSO, each particle represents its position by either the binary value {0} or {1}, and the velocity is treated as a probability change of the particle position. However, BPSO has the same disadvantage as other evolutionary algorithms. After several generations, these algorithms tend to easy get trapped in a local optimum, which might prevent them from converging towards a global optimal solution. In order to circumvent the premature convergence at a local optimum, we incorporated a simple Boolean operation to create a new *gBest* position. This new *gBest* replaced the original *gBest*, so that all particles were able to leave the local optimal.

In this study, we compared the gene selection performance of the filter and wrapper models, and hybrid the two models to create a new hybrid model for gene selection. To evaluate and compare the proposed method to other feature selection methods, we used two the K-nearest neighbor (KNN) and a Support Vector Machine (SVM) classification algorithm to evaluate the selected features, and to establish the influence on classification accuracy. The results indicate that in terms of the number of genes that need to be selected and

L. Y. Chuang is with the Chemical Engineering Department, I-Shou University, 84001, Kaohsiung, Taiwan. (E-mail: chuang@isu.edu.tw).
C.H. Yang and C. H. Ke are with the Electronic Engineering Department, National Kaohsiung University of Applied Sciences, 80778, Kaohsiung, Taiwan. (phone: 886-7-3814526#5639; e-mail: chyang@cc.kuas.edu.tw, a092205237@cc.kuas.edu.tw).

classification accuracy the proposed method is superior to other methods in the literature. This paper is organized as follows: a brief overview introducing the methods is presented in Section II. The experimental framework and settings are described in Section III. Section IV consists of the results and a theoretical discussion thereof. Finally, the concluding remarks are offered in Section V.

## II. RELATED METHODS

### A. Information Gain

Quinlan [5] proposed a classification algorithm called ID3, which introduces the concept of information gain. Information gain is a measure based method, which is usually used to select best split attributes in decision tree classifiers. The measure indicates to what extent the entire data's entropy is reduced, and identifies the value of each specific attribute. Each feature basis obtains an information gain value, the amount of which is used to decide whether the feature is selected or deleted. Therefore a threshold value for selecting a feature must first established; a feature is selected when the information gain value of this feature is bigger than the threshold value.

Let S be the set of n instances and let C be the set of k classes. Let $P(C_i, S)$ be the fraction of the examples in S that have class $C_i$, Then, the expected information from this class membership is as follows:

$$Info(S) = -\sum_{i=1}^{k} P(C_i,S) \times \log(P(C_i,S)) \quad (1)$$

If a particular attribute A has v distinct values, the expected information required for the decision tree with A as the root is then the weighted sum of expected information of the subsets of A according to distinct values. Let $S_i$ be the set of instances whose value of attribute A is $A_i$.

$$Info_A(S) = -\sum_{i=1}^{v} \frac{|S_i|}{|S|} \times Info(S_i) \quad (2)$$

Then, the difference between Info(S) and $Info_A(S)$ gives the information gained by partitioning S according to testing A.

$$Gain(A) = Info(S) - Info_A(S) \quad (3)$$

The higher the information gain, the higher are the chances of getting pure classes in a target class if split on the variable with the highest gain.

### B. Continuous particle swarm optimization

The PSO system is initialized with a population of random solutions. This population searches for an optimal solution by updating generations. In PSO, a potential solution is called a particle. Each particle makes use of its own memory and knowledge gained by the swarm as a whole to find the best (optimal) solution in a *d*-dimensional search space. The particles have a positional value and velocities which direct their movement.

Each particle is represented by $x_i = (x_{i1}, x_{i2}, ..., x_{id})$, where *d* is the dimension numbers. The rate of velocity for the $i_{th}$ particle is represented by $v_i = (v_{i1}, v_{i2}, ..., v_{id})$ and limited by $V_{max}$, which is determined by the user. The best previously encountered position of the $i_{th}$ particle (the position with the highest fitness value) is called $pBest_i$ and represented by $p_i = (p_{i1}, p_{i2}, ..., p_{id})$. The global best value of the entire population is called *gBest* and represented by $g = (g_1, g_2, ..., g_d)$. At each interaction, the particles are updated according to the following equations:

$$v_{id}^{new} = w \times v_{id}^{old} + c_1 \times rand_1 \times (pbest_{id} - x_i^{old})$$
$$+ c_2 \times rand_2 \times (gbest_d - x_i^{old}) \quad (4)$$

$$x_{id}^{new} = x_{id}^{old} + v_{id}^{new} \quad (5)$$

where w is the inertia weight, $c_1$ and $c_2$ are acceleration (learning) factors, $rand_1$ and $rand_2$ are random numbers. Velocities $v_{id}^{new}$ and $v_{id}^{old}$ are those of the new and old particle, respectively, $x_{id}^{old}$ is the current particle position (solution), and $x_{id}^{new}$ is the updated particle position.

### C. Binary Particle Swarm Optimization

Although PSO was originally introduced as an optimization technique for real-number optimization problems, many optimization problems are set in a space featuring discrete or qualitative distinctions between variables. PSO and BPSO each have their own characteristics. In the first, each particle is composed of a binary variable. Each particle will decide on "yes" or "no", "true" or "false" or {1} or {0} in this model. In the latter however, the velocity is transformed into a change of probability, namely the probability of the binary variable taking the value {1}. However, the velocity must be restricted to the range [0.0, 1.0]. In order to map the real value number of the velocity to the range, the sigmoid function is proposed to handle the probability of the variables [3].

$$S(v_{pd}^{new}) = \frac{1}{1+e^{-v_{pd}^{new}}} \quad (6)$$

where $S(v_{pd}^{new})$ denotes the probability of bit $x_{pd}^{new}$, if $(rand() < S(v_{pd}^{new}))$ then $x_{pd}^{new} = 1$; else $x_{pd}^{new} = 0$, the *rand()* is a random number selected from a uniform distribution in [0.0, 1.0]. To avoid $S(v_{pd}^{new})$ approaching 0 or 1, a constant $V_{max}$ is used to limit the $v_{pd}^{new}$, the range of maximum velocity is $[-V_{max}, +V_{max}]$. $V_{max}$ is usually set to 6.

### D. An improved Binary Particle Swarm Optimization

In this study, we propose an improved binary particle swarm optimization (IBPSO) which further develops standard BPSO. We executed IBPSO for feature selection and compared it to traditional BPSO. In general, after several generations, particles are influenced by their own *pBest* value and will stop moving towards *gBest*. If *gBest* is not changed, the particles cluster around *gBest*. To prevent particles from getting trapped in a local optimum, we introduced a simple Boolean algebra operation.

The primitive Boolean function has three simple logical operators, namely '*and*' (·), '*or*' (+), and '*not*' (−). The three operations can combine all kinds of logical functions and are used to solve combinational logic problems. In IBPSO, we assume that if the *gBest* values are unchanged after threegenerations, the particles have fallen into the local optimum. The particles thus stuck have to be induced to leave the local optimum. We used the '*and*' (·) logical operation to '*and*' *pBest* of all particles. After the operation, a new binary string will be created, and be treated as the new *gBest*. After this new *gBest* is created, each particle will depart from its original position, and continue to search other place in the search space.

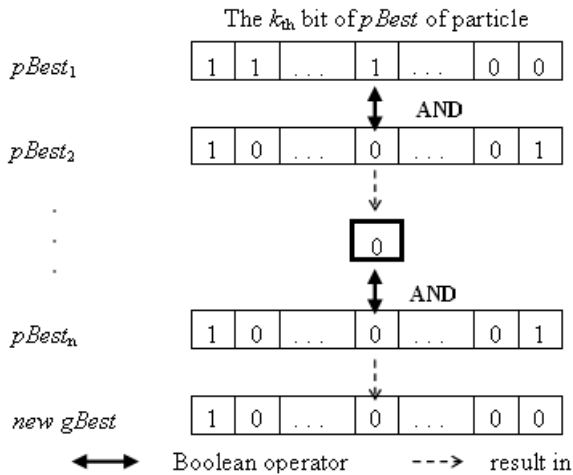

Figure 1. Process of creating a new *gBest*

III. EXPERIMENT FRAMEWORK

*A. Hybrid filter and wrapper feature selection method*

In this study, we hybrid the filter and wrapper model methods to select feature genes in microarrays, and used two different classification algorithms to evaluate the performance of the proposed method. Figure 2 depicts the process of the hybrid filter and wrapper model feature selection method. The filter model part uses information gain (IG) to evaluate the ability of each feature which differentiates between different categories. The reasoning behind this method is that it can calculate the importance of each feature with respect to the class. We used Weka [5] to determine the information value of each feature and sort the features in accordance with their information gain value. Higher values indicate higher discrimination of this feature from other categories, and mean that this feature can be used to calculate classification results effectively. After calculating the information gain values of all features, we implemented a threshold for the results. Since after calculation most information gain values were zero, not many features have an influence on the categorization of a data set. The threshold in our study is 0 for most data sets. If the information gain values of the features are higher than the threshold, we select the feature, if not, the feature is not selected.

For example, let a microarray data set have 10 gene numbers (10 feature numbers which can be represented by $f_1 f_2 f_3 f_4 f_5 f_6 f_7 f_8 f_9 f_{10}$). If only 5 genes ($f_1$, $f_2$, $f_4$, $f_7$ and $f_{10}$) conform to the information gain threshold, only these 5 genes ($f_1 f_2 f_4 f_7 f_{10}$) are used after the wrapper procedure to implement the selection process.

However, when using the filter model selection, the feature number could be reduced dramatically. In order to more effectively remove unwanted features, we used traditional BPSO and the improved BPSO method for wrapper model selection after the initial filter model selection was implemented to select features again, and used KNN and SVM algorithm to measure the classification performance.

In BPSO, we used two bits, {0} and {1}, to represent the feature condition. If the number of each particle is identical to the input data dimension number, the $i_{th}$ particle is $p_i = 10011$ in our example, meaning that the $f_1$, $f_7$ and $f_{10}$ feature are selected, and the others are not select.

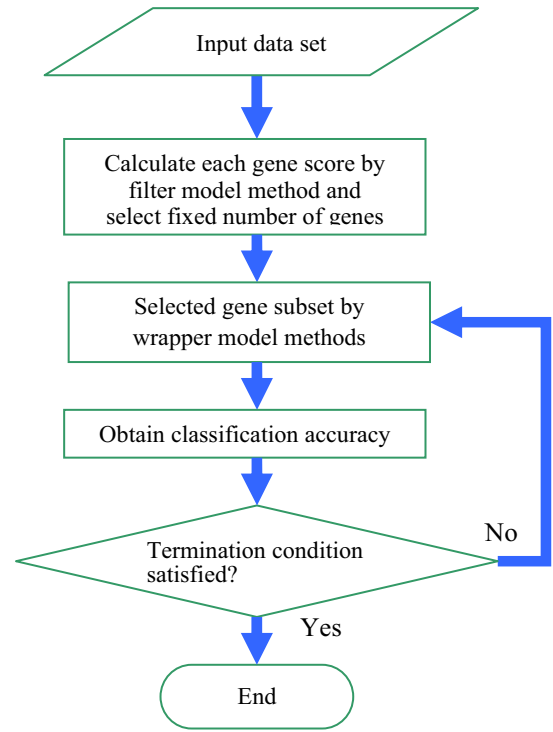

Figure 2. Hybrid filter and wrapper model feature selection method

*B. Data sets*

We used six multi-category cancer-related human gene expression data sets [4], which were downloaded from http://www.gems-system.org to evaluate the performance of the proposed method. The data format is shown in Table 1; it includes the data set name, the number of samples, categories and samples. In order to avoid bias, we implemented a linear scale for each gene expression data value to the range [0, 1].

*C. Classifier performance*

To evaluate the performance of the proposed method, the selected feature subsets were evaluated by leave-one-out cross-validation (LOOCV) of one nearest neighbor (1-NN), and K-fold cross validation (K-fold) for SVM.

For LOOCV, neighbors are calculated using their Euclidean distance. The fitness value for the 1-NN evolves according to the LOOCV method for all data sets. In the LOOCV method, a single observation from the original sample is selected as the validation data, and the remaining observations are selected as training data. This is repeated so that each observation in the sample is used once as the validation data. For K-fold cross validation, we set K=10 in this study. During K-fold cross-validation, the data was separated into 10 parts $\{D_1, D_2, \ldots, D_{10}\}$, and training and testing was carried out a total of 10 times. When any part $D_n$, $n = 1, 2, \ldots, 10$ is processed as a test set, the other 9 parts will be training sets. Following 10 times of training and testing, 10 classification accuracies are produced, and the averages of these 10 accuracies are used as the classification accuracy for the data set. We assumed that the obtained classification accuracy is an adaptive functional value.

Furthermore, the one-versus-rest method (OVR) was used to deal with multi-class problems. OVR assembles classifiers that distinguish one class from all other classes. For each class i, $1 \leq i \leq k$, a binary classifier separating class i from the rest is built. To predict a class label of a given data point, the output of each of the k classifiers is obtained. If there is a unique class label, say j, which is consistent with all the k predictions, the data point is assigned to class j. Otherwise; one of the k classes is selected randomly. Very often though, a situation arises in which consistent class assignment does not exist.

*D. Experimental parameter set*

Principal algorithm parameter settings used were the following. The number of particles used was 30. The two factors rand1 and rand2 are random numbers between [0, 1], whereas c1 and c2 are acceleration factors, here c1 = c2 = 2. The inertia weight w was 1.0. The maximum number of iterations was 100. For the GA, chromosomes numbers is 30, crossover rate, and mutation rate is 1.0 and 0.1 respectively.

IV. RESULTS AND DISCUSSION

In this study, we tested and compared a hybrid filter and wrapper feature selection method's performance on the classification of six multi-category cancer microarray expression data sets. We also proposed an improved BPSO method and compared it to other evolutionary algorithms. After feature selection, the selected feature subsets were evaluated using two common classification algorithms.

Table 2 and Table 3 show the accuracies achieved by the filter, wrapper and hybrid model feature selection methods individually. In Table 2, the classification accuracy is evaluated by KNN and in Table 3 by SVM. The experimental results show that the accuracy of microarray data which had feature selection implemented was better than without feature selection. For all the feature selection methods, the average of the wrapper model accuracy was better than for the filter model, but the number of selected feature was also higher for the wrapper model than for the filter model. The wrapper model differs from the filter model in that it is dependent on a classifier and evaluates the combination of feature subsets. The wrapper model can identify interaction amongst all features simultaneously.

However, how many gene subsets are truly necessary to identify cancer categories is still a question under debate [8]. Only filter selection does not reduce the number of features very much; hence another method is needed to reduce the number of features further. In order to select more effective feature subsets, we acceded to a wrapper model method after the implementing the filter approach. Table 2 and Table 3 show that the proposed method effectively increases classification accuracy and selects a smaller number of feature subsets.

During the wrapper phase of the proposed method, we implemented an improved BPSO algorithm, which uses a Boolean function to prevent the standard BPSO's premature convergence on a local optimal, and compared it to a GA and BPSO. The experiment showed that the combination of IBPSO and a wrapper model alone or when yet hybrid with a filter method achieves a better performance than GA or single BPSO. The reason for this is simple: all evolutionary algorithms are prone to converge at a local optimum location. In a GA, for example, chromosomes with higher fitness values are chosen as parents and produce a new offspring. After several generations (crossover and mutation), all chromosomes are similar, which reduces the search capability. In BPSO, each particle is guided by *pBest* and *gBest*. However, if *gBest* is not continuously changed, all of the particles will close in on *gBest* after several generations, again resulting in a weakened search capability.

V. CONCLUSION

In this paper, we hybrid the filter and wrapper model methods for microarray classification to implement a feature selection process, and then used KNN and SVM to evaluate the classification performance. Experimental results showed that the proposed method simplified gene selection and the total number of parameters needed effectively, thereby obtaining a higher classification accuracy compared to other feature selection methods. The classification accuracy obtained by the proposed method was higher than other methods for all six test problems. In the future, the proposed method can assist in further research where feature selection needs to be implemented. It can potentially be applied to problems in other areas as well.

Table 1. Cancer-related human gene expression datasets

| Dataset Name | Diagnostic task | Number of Sample | Number of Genes | Number of Classes |
|---|---|---|---|---|
| 9_Tumors | Nine various human Tumor types | 60 | 5726 | 9 |
| Brain_Tumor1 | Five human brain tumor types | 90 | 5920 | 5 |
| Brain_Tumor2 | Four malignant glioma types | 50 | 10367 | 4 |
| Leukemia1 | Acute myelogenous leukemia (AML), acute lympboblastic leukemia (ALL) B-cell, and ALL T-cell | 72 | 5327 | 3 |
| Leukemia2 | AML, ALL, and mixed-lineage leukemia (MLL) | 72 | 11225 | 3 |
| DLBCL | Diffuse large B-cell lymphomas and follicular lymphomas | 77 | 5469 | 2 |

Table 2. KNN accuracy performance for the six microarray data sets for the filter, wrapper and hybrid filter/wrapper feature selection methods

| Data set | KNN (Statnikov et al.)[1] | filter IG | wrapper GA | wrapper BPSO | wrapper IBPSO | Hybrid IG + GA | Hybrid IG + BPSO | Hybrid IG + IBPSO |
|---|---|---|---|---|---|---|---|---|
| 9_Tumors | 43.90 | 66.67 | 56.67 | 60.00 | 70.00 | 85.00 | 85.00 | 90.00 |
| Brain_Tumor1 | 87.94 | 88.89 | 92.22 | 91.11 | 93.33 | 93.33 | 93.33 | 96.67 |
| Brain_Tumor2 | 68.67 | 78.00 | 80.00 | 80.00 | 86.00 | 88.00 | 84.00 | 92.00 |
| Leukemia1 | 83.57 | 93.06 | 97.22 | 94.44 | 97.22 | 100.0 | 98.61 | 100.0 |
| Leukemia2 | 87.14 | 91.67 | 95.83 | 91.67 | 97.22 | 98.61 | 95.83 | 100.0 |
| DLBCL | 86.96 | 93.51 | 93.51 | 90.91 | 96.10 | 100.0 | 100.0 | 100.0 |
| Average | 76.36 | 85.30 | 85.91 | 84.69 | 89.98 | 94.16 | 92.80 | 96.45 |

Table 3. SVM accuracy performance for the six microarray data sets for the filter, wrapper and hybrid filter/wrapper feature selection methods

| Data set | MC-SVM (No FS) (Statnikov et al.)[1] | MC-SVM (with FS) (Statnikov et al.)[1] | filter IG | wrapper GA | wrapper BPSO | wrapper IBPSO | Hybrid IG + GA | Hybrid IG + BPSO | Hybrid IG + IBPSO |
|---|---|---|---|---|---|---|---|---|---|
| 9_Tumors | 65.10 | 74.86 | 75.00 | 71.67 | 71.67 | 78.33 | 90.00 | 90.00 | 90.00 |
| Brain_Tumor1 | 91.67 | 92.67 | 91.11 | 90.00 | 90.00 | 90.00 | 91.11 | 92.22 | 92.22 |
| Brain_Tumor2 | 77.00 | 85.67 | 84.00 | 84.00 | 80.00 | 88.00 | 88.00 | 88.00 | 88.00 |
| Leukemia1 | 97.50 | | 95.71 | 97.14 | 97.14 | 98.57 | 97.14 | 97.14 | 98.57 |
| Leukemia2 | 97.32 | | 97.14 | 98.57 | 98.57 | 100.0 | 98.57 | 98.57 | 98.57 |
| DLBCL | 97.50 | | 97.14 | 98.57 | 97.14 | 98.57 | 97.14 | 98.57 | 98.57 |
| Average | 87.68 | | 90.02 | 89.99 | 89.09 | 92.25 | 93.66 | 94.08 | 94.32 |

Table 4. The selected feature number for the six microarray data sets the filter, wrapper and hybrid filter/wrapper feature selection methods using SVM

| Data set | MC-SVM (No FS) (Statnikov et al.)[1] | MC-SVM (with FS) (Statnikov et al.)[1] | filter IG | wrapper GA | wrapper BPSO | wrapper IBPSO | Hybrid IG + GA | Hybrid IG + BPSO | Hybrid IG + IBPSO |
|---|---|---|---|---|---|---|---|---|---|
| 9_Tumors | 5726 | 500 | 165 | 1171 | 1531 | 552 | 52 | 49 | 31 |
| Brain_Tumor1 | 5920 | 500 | 1612 | 954 | 2602 | 435 | 244 | 474 | 115 |
| Brain_Tumor2 | 10367 | 500 | 4465 | 1018 | 4158 | 1081 | 489 | 1855 | 327 |
| Leukemia1 | 5327 | | 848 | 974 | 1811 | 462 | 82 | 186 | 34 |
| Leukemia2 | 11225 | | 4596 | 2002 | 4310 | 829 | 782 | 459 | 220 |
| DLBCL | 5469 | | 882 | 1244 | 2123 | 485 | 107 | 252 | 33 |

**Legends:** (1) MC-SVM: Multi-class support vector machines (2) FS: Feature selection (3) IG: Information Gain (4) GA: Genetic Algorithm (5) BPSO: Binary Particle Swarm Optimization (6) IBPSO: Improved Binary Particle Swarm Optimization